\documentclass[11pt]{article}

\usepackage[preprint]{acl}

\usepackage{times}
\usepackage{latexsym}
\usepackage[T1]{fontenc}
\usepackage[utf8]{inputenc}
\usepackage{microtype}
\usepackage{inconsolata}

\usepackage{graphicx}
\usepackage{booktabs}
\usepackage{amsmath}
\usepackage{amssymb}
\usepackage{algorithm}
\usepackage{algorithmic}
\usepackage[most]{tcolorbox}
\usepackage{xcolor}
\usepackage{placeins}
\usepackage{cleveref}

\definecolor{lightgray}{gray}{0.96}
\tcbuselibrary{listingsutf8}
\tcbset{
  mysinglecolumnbox/.style={
    colback=lightgray,
    colframe=black!40,
    arc=2mm,
    boxrule=0.4pt,
    fontupper=\ttfamily\footnotesize,
    left=2mm, right=2mm, top=1mm, bottom=1mm,
    enhanced,
    breakable,
    before upper={\sloppy\emergencystretch=3em\relax}
  },
  mycodebox/.style={
    colback=white,
    colframe=black!20,
    boxrule=0.3pt,
    fontupper=\ttfamily\scriptsize,
    arc=1mm,
    left=1mm, right=1mm, top=1mm, bottom=1mm,
    enhanced,
    breakable,
    before upper={\sloppy\emergencystretch=3em\relax}
  }
}

\title{{HDSL}: A Hierarchical Domain-Specific Language for Structured 3D Indoor Scene Generation and Localized Editing with LLM Agents}

\author{
\begin{tabular}{c}
Letian Li$^{1}$, Chao Shen$^{2}$, Shuzhao Xie$^{1}$, Chenghao Gu$^{1}$, ZhengXiao He$^{3}$, \\
Yu Meng$^{1}$, Xin Yang$^{4}$, Wenyuan Jiang$^{5}$, Zhi Wang$^{1\dagger}$ \\
\textnormal{\normalsize
$^1$SIGS, Tsinghua University \quad
$^2$Nankai University \quad
$^3$University of Arizona} \\
\textnormal{\normalsize
$^4$Zhejiang University \quad
$^5$ETH Zurich} \\
\texttt{\normalsize lilt24@mails.tsinghua.edu.cn, wangzhi@sz.tsinghua.edu.cn}
\end{tabular}
}

\begin{document}
\maketitle

\begin{abstract}
Text-driven indoor scene generation and editing require an intermediate representation that language models can both produce and revise. Existing LLM-based systems often rely on scene graphs or global constraint lists, which are compact but underspecify local geometry and make instruction-based edits difficult to localize. We frame this problem as structured program generation and local program repair, and propose Hierarchical Descriptive Scene Language (HDSL), an XML/CSS-style domain-specific language for structured 3D indoor scenes. HDSL represents rooms, regions, objects, and support surfaces as a tree with local coordinates, making complex scenes easier to plan recursively and easier to retrieve for editing. Our pipeline uses LLM agents to generate HDSL subtrees with bounded verification, grounds non-virtual nodes through multimodal asset retrieval, and applies force-directed layout optimization to repair boundary and collision errors. For editing, Hierarchical Retrieval-Augmented Generation retrieves the relevant subtree, asks the LLM to rewrite only that local context, and merges the result back through a deterministic three-way merge. In our reproduced benchmark, HDSL improves average object coverage, text-scene alignment, and generation time over full text-to-scene baselines while remaining competitive with recent layout-only reproductions on geometry metrics; for editing, HRAG reduces token use by $5.22\times$ and runtime by $6.19\times$, produces valid DSL for all eight paired edits, and better preserves unrelated scene objects.
\end{abstract}

\section{Introduction}
\label{sec:intro}

\begin{figure}[t]
    \centering
    \includegraphics[width=\columnwidth]{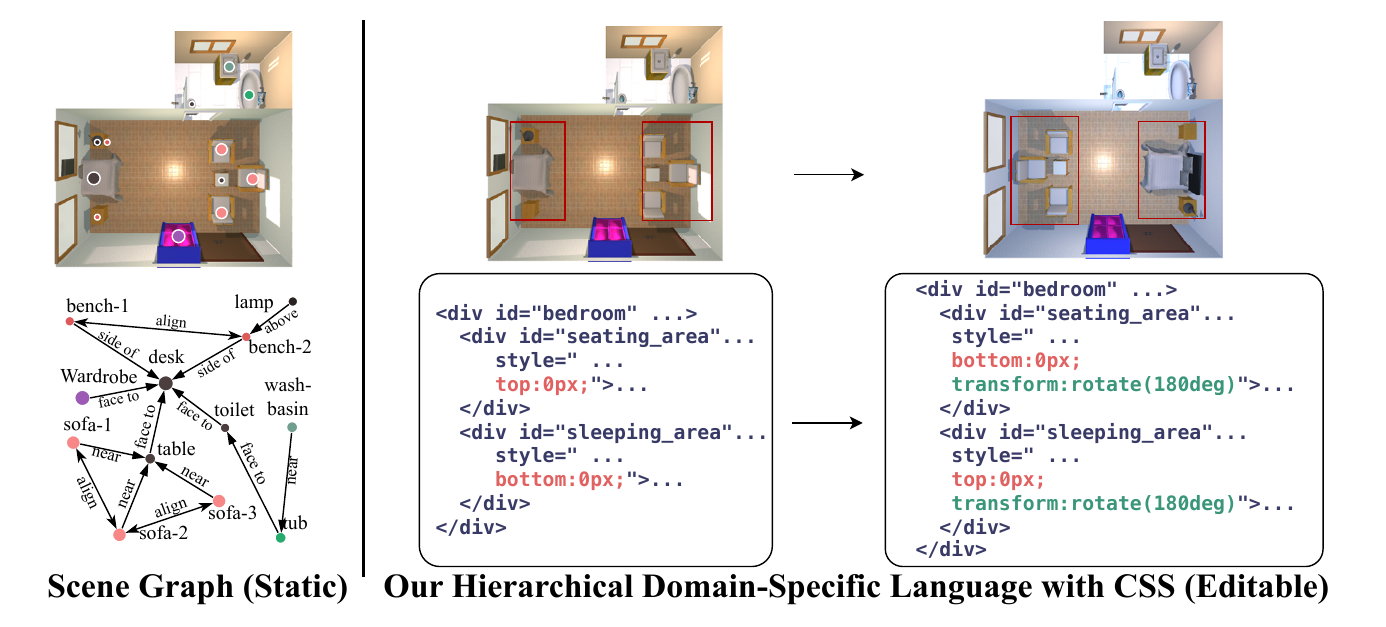}
    \caption{\textbf{HDSL as an editable scene interface.} Compared with flat scene graphs, an XML/CSS-style hierarchy stores local geometry, support relations, and object identity in addressable nodes, so an edit can target the relevant subtree instead of rewriting the whole scene.}
    \label{fig:teaser} 
\end{figure}

Text-driven indoor scene generation and editing asks a system to convert a natural-language request into a renderable 3D indoor scene, and then revise that scene through follow-up instructions. This task supports VR~\cite{luong2022vr}, AR~\cite{suzuki2022ar}, embodied AI~\cite{duan2022surveyembodiedaisimulators}, and rapid design prototyping, where users need both initial scene construction and iterative modification. Agentic LLM pipelines are especially promising here: instead of training a separate generator for every domain, an agent can decompose a prompt into planning steps, retrieve assets, call geometric checks, and revise intermediate outputs through tool feedback.

The current agentic route, however, remains expensive and hard to control. Holodeck~\cite{Yang_2024_CVPR_Holodeck} and I-Design~\cite{celen2024idesign} ask LLMs to produce scene graphs or constraint sets before retrieving and arranging Objaverse-scale assets~\cite{deitke2023objaverse}; DirectLayout~\cite{ran2025direct} and LayoutVLM~\cite{sun2025layoutvlm} further explore language-driven layout construction. These systems demonstrate the flexibility of agents, but their scene interfaces often leave metric geometry and object identity distributed across solvers, validators, and post-processing state. As a result, a complex room may require long prompts, repeated retries, and global checks even when the user wants a small local change. Editing makes this cost more visible: follow-up instructions often require feeding back a large scene context or rerunning a broad generation process, which increases token use and raises the risk of moving, deleting, or re-grounding unrelated objects~\cite{boudjoghra2025scanedit}.

These problems are difficult because indoor scenes are nested, support-heavy, and partly metric. A room contains functional zones, zones contain furniture groups, and many objects depend on tables, shelves, counters, or walls. A representation for an LLM therefore has to expose local coordinates, support relations, and persistent object identity while keeping each prompt small enough to reason over reliably. Prior work often treats this as a downstream repair problem, adding stronger validators, backtracking, render feedback, or ad hoc edit rules after the scene interface has already been chosen. These components are useful, but they do not make token consumption and localized editing the central design target.

We instead treat the scene representation as the control surface for the agent. Inspired by LLM-based web development~\cite{xiao2025designbench} and CSS layout generation~\cite{feng2024layoutgpt}, we introduce \textbf{Hierarchical Descriptive Scene Language (HDSL)}, an XML/CSS-style DSL for 3D indoor scenes. HDSL acts as a harness around the LLM: it constrains outputs to parseable nodes, stores local geometry in the parent frame, exposes deterministic checks, and gives each room, region, object, and support container an addressable identity, as shown in~\cref{fig:teaser}. Our pipeline generates HDSL subtrees recursively, grounds concrete nodes through multimodal asset retrieval, repairs residual spatial errors with force-directed layout optimization, and edits scenes by retrieving and rewriting only the relevant subtree before a deterministic three-way merge. Our contributions are:

\begin{itemize}
    \item We propose \textbf{HDSL}, a hierarchical XML/CSS-style scene DSL that turns 3D indoor generation into bounded, checkable, and addressable structured prediction for LLM agents.
    \item We design a \textbf{hierarchical generation pipeline} with recursive LLM agents, sibling-level coordination, generate--verify--revise checks, multimodal asset grounding, and force-directed layout optimization.
    \item We introduce \textbf{HRAG-HDSL} for localized instruction-based editing and validate the representation on generation, editing, ablation, and user-study experiments; HRAG reduces token use by $5.22\times$ and runtime by $6.19\times$ while better preserving non-target objects than full-scene rewriting.
\end{itemize}

\section{Related Work}
\label{sec:related_work}

\vspace{1mm}\noindent\textbf{Structured LLM Interfaces and DSLs}. Many LLM systems ask the model to produce structured intermediate forms rather than free text. Declarative prompting can hand symbolic constraints to a solver~\cite{ye2023SatLM}, and program-generation approaches can move arithmetic or spatial reasoning into an external language or interpreter~\cite{yang2024arithmetic,aguinakang2024openuniverse}. These methods work because the output format constrains the model and exposes errors to deterministic checks. For indoor scene generation, however, the intermediate form must also carry support relations, local coordinates, and object identity so that later instructions can modify only the affected part of the scene.

Current LLM-based scene generators such as Holodeck~\cite{Yang_2024_CVPR_Holodeck} and I-Design~\cite{celen2024idesign} use scene graphs or constraint lists as the intermediate representation. These representations are compact for one-shot generation, but they do not expose a stable, addressable program surface for later edits. HDSL instead uses an XML/CSS-style representation: it keeps the syntax close to familiar web layout, but adds scene-specific attributes for 3D geometry, support relations, and retrieval metadata for downstream grounding.

\vspace{1mm}\noindent\textbf{Scene Representation}. Indoor scene generation uses both 3D representations and text-facing representations. Point clouds and voxel grids are common for learned end-to-end models, while LLM-based systems often use scene graphs, constraint lists, or executable programs before asset retrieval and placement. Scene Language~\cite{zhang2024SceneLanguage}, for example, uses a Lisp-like code form and an interpreter to translate symbolic scene code into a final representation.

For generation with follow-up editing, the representation must do more than describe a scene: it must let the system locate the affected region, preserve unrelated content, and expose enough geometry for a local rewrite. Scene graphs are often too abstract for this purpose, while procedural programs can hide the final state behind control flow. HDSL uses a static declarative tree, so every generated object remains directly indexable and editable.

\vspace{1mm}\noindent\textbf{Learning-based Methods}. Learned indoor-scene generators usually train directly on 3D scene data. DiffuScene~\cite{tang2024diffuscene} and PhyScene~\cite{yang2024physcene} use layout diffusion for single-room generation. End-to-end diffusion methods, such as DiffInDScene~\cite{ju2024diffindscene} and LT3SD~\cite{meng2024lt3sdlatenttrees3d}, synthesize finer-grained scenes from annotated 3D data.

However, learning-based scene generation models often need costly training and large datasets, making them difficult to apply to scene types with limited examples, such as industrial warehouses. Our approach does not train a new scene generator; it designs a recursive interface that lets a pretrained LLM plan unseen scene types through local, verifiable layout decisions.

\vspace{1mm}\noindent\textbf{LLM-based Methods}. LLM-based scene generation methods can be broadly categorized by how they represent and resolve layouts. Constraint-based approaches such as Holodeck~\cite{Yang_2024_CVPR_Holodeck} and I-Design~\cite{celen2024idesign} have the LLM emit a set of spatial constraints, which are then solved by a search or backtracking algorithm. DirectLayout~\cite{ran2025direct} and LayoutVLM~\cite{sun2025layoutvlm} further study prompt-driven or vision-language layout optimization, but their public pipelines are most reproducible as layout-only systems with fixed or externally supplied object sets. HSM~\cite{pun2026hsm} introduces hierarchical scene motifs over HSSD-HAB assets, but its full pipeline depends on support-surface annotations and expensive VLM feedback. These systems improve different parts of the generation pipeline, yet they still provide limited support for partial scene modification without re-solving or re-generating a large context. Our method treats the scene representation itself as the model interface: each node is spatially self-contained within its parent, so relevant subtrees can be retrieved, edited, and merged back without changing the rest of the global scene state.

\section{Methods}

\begin{figure*}[t]
    \centering
    \includegraphics[width=\textwidth]{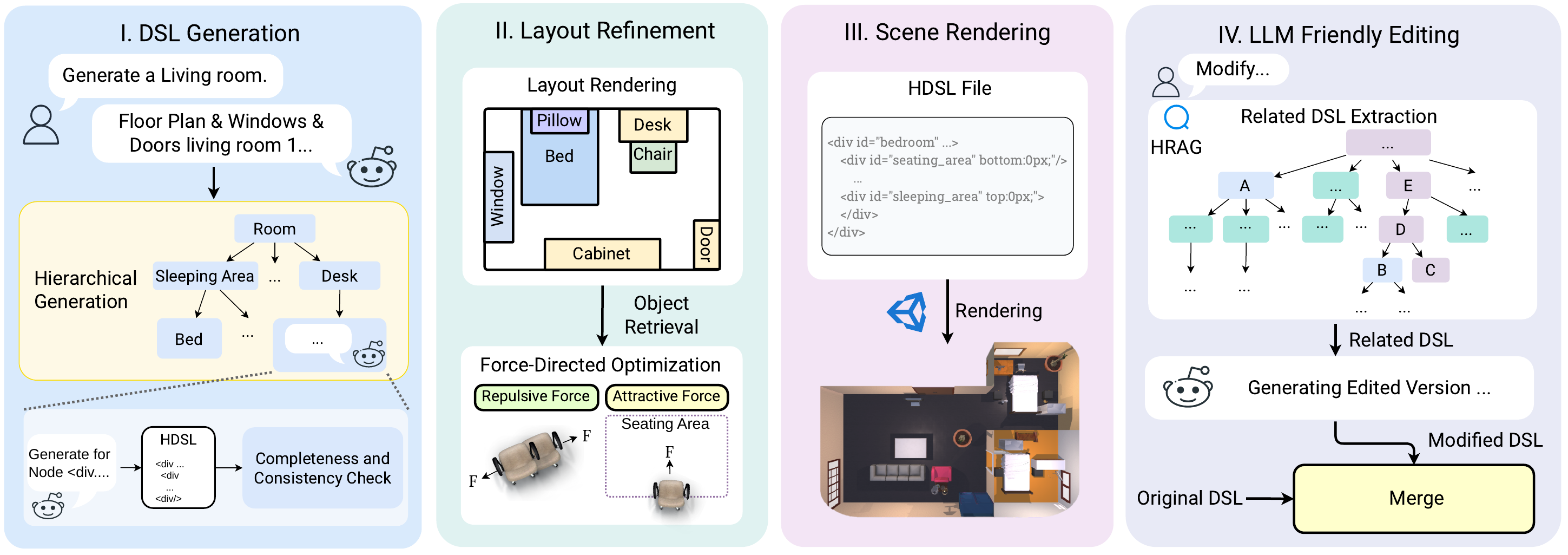}
    \caption{\textbf{HDSL framework} for scene generation and editing proceeds in four stages:  \textbf{I. DSL Generation}. Given a natural language prompt, recursively constructing the scene structure using LLM agents.  \textbf{II. Layout Refinement}. Parsing the CSS, retrieving assets, and optimizing object placements via force-directed layout to eliminate overlaps and boundary violations. \textbf{III. Rendering}. Composing assets based on the DSL and scene floorplan config. \textbf{IV. LLM Friendly Editing}. Given a user editing prompt, retrieving relevant subtrees for localized modifications via LLM agents, and reintegrating them into the DSL.  }
    \label{fig:method}
\end{figure*}

We view text-to-scene generation as structured prediction from a natural-language prompt $P$ to a scene program $\Phi(\mathcal{S})$, and instruction-based editing as program repair from $(\Phi(\mathcal{S}), P_{\mathrm{edit}})$ to an updated program $\Phi'(\mathcal{S})$. This formulation makes the interface between the LLM and the scene renderer explicit: the model writes and revises a parseable program, while deterministic modules validate geometry, retrieve assets, and merge local edits.

Our method is built around three requirements: a scene representation that exposes local geometry to the LLM, a generation procedure that plans complex rooms through bounded local decisions, and an editing procedure that changes only the relevant subtree, as shown in~\cref{fig:method}.

\subsection{Hierarchical Descriptive Scene Language}

\begin{figure}[t]
  \centering
  \includegraphics[width=\columnwidth]{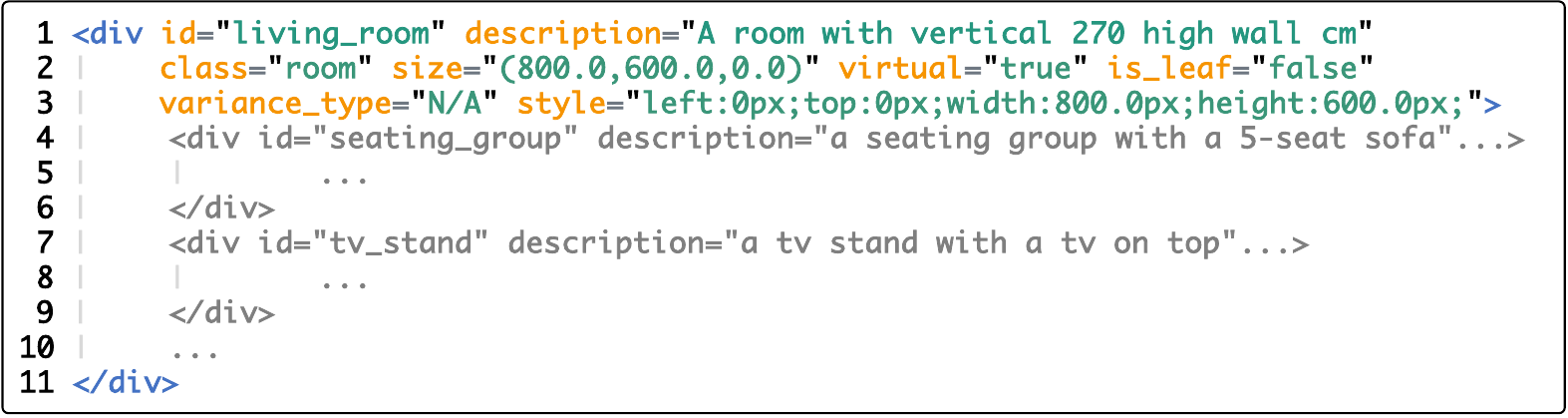}
  \caption{\textbf{Example HDSL of a living room}. The snippet shows nested virtual zones, concrete objects, support relations, and local CSS-style spatial attributes. Due to space constraints, we show only the essential code needed to illustrate the hierarchy.}
  \label{fig:spec}
\end{figure}

HDSL represents an indoor scene as an XML/CSS tree. Formally, a scene $\mathcal{S}$ is encoded as $\Phi(\mathcal{S})=(\mathcal{N},\mathcal{E})$, where each node $n\in\mathcal{N}$ denotes a room, region, object, or virtual support container, and each edge in $\mathcal{E}$ denotes a parent-child containment or support relation. Each node stores semantic, geometric, hierarchy-related, and retrieval-related attributes; the full schema and an example are provided in~\cref{tab:hdslasspec,fig:spec}.

This representation is both a generation target and an editing index. CSS supplies local coordinates and rotation in the parent frame, while XML nesting keeps the context short and directly addressable. Unlike a flat JSON list or a symbolic scene graph, the parent-child syntax itself records containment, support, and local coordinate frames, so retrieval and editing can operate on subtrees rather than on a reconstructed global state. The tree also matches indoor structure: rooms decompose into functional zones, objects form local groups, and many small objects are placed on supporting surfaces. See Sec.~\ref{sec:hdsl_lifting} for the full lifting rule.

\subsection{Hierarchical Generation via LLM Agents}

HDSL turns generation into local recursive planning rather than asking the LLM to instantiate the whole scene at once. For each expandable node $n_i$, the LLM receives its ancestor path $\mathcal{P}(n_i)=\{n_0=r,\dots,n_k\}$, parent geometry, inherited floorplan information, global instructions, and few-shot examples~\cite{dong-etal-2024-survey}. This makes each prompt a bounded structured-output task: the model only needs to emit the direct children of one container while preserving enough scene context for functional consistency.

At each step, the LLM decides the child set $\mathcal{C}(n_i)$ and emits, for each child $c_j$, its semantic description, class, 3D size, virtual flag, leaf flag, and CSS pose. The virtual flag distinguishes spatial regions or support containers from objects that require asset retrieval, while the leaf flag determines whether the child should be recursively expanded in later iterations. To avoid sparse or overcrowded local layouts, we score the occupancy of the generated children:
$$
\rho(n_i)=\frac{\sum_{c_j\in\mathcal{C}(n_i)}\mathrm{area}(c_j)}{\mathrm{area}(n_i)}
$$
together with validity and semantic coverage. We use a bounded multi-candidate generate--verify--revise loop: candidates are parsed as XML, checked for structural and spatial validity, repaired when deterministic fixes are safe, and otherwise rejected with feedback for the next attempt. Siblings are generated in one call so their relative positions are coordinated, while completed disjoint subtrees can be expanded in parallel. Additional prompt constraints, validators, and thresholds are provided in \cref{sec:llm_sanity}.

\vspace{1mm}\noindent\textbf{Object Retrieval.} After HDSL generation, non-virtual nodes are grounded with Objaverse assets~\cite{deitke2023objaverse} using the Holodeck retrieval pipeline~\cite{Yang_2024_CVPR_Holodeck} and OpenCLIP matching~\cite{openclip}. Retrieval details are provided in the appendix.

\subsection{Force-Directed Layout Optimization}

LLM-generated layouts may still contain boundary violations, sibling overlaps, overly tight virtual containers, or wall-facing objects. We therefore apply a recursive force-directed refinement within each parent node. For child $i$, FDLO combines a boundary attraction term and an overlap repulsion term:
$$
\vec{F}_{\mathrm{a}}=\alpha \frac{\vec{D}_i}{\|\vec{D}_i\|}, \quad
\vec{F}_{\mathrm{r}}^{ij}=\beta\,\mathrm{IoU}_{ij}\frac{\vec{\Delta}_{ij}}{\|\vec{\Delta}_{ij}\|}.
$$
Here, $\vec{D}_i$ points from the child center toward the valid parent region, $\vec{\Delta}_{ij}$ separates overlapping siblings, and $\mathrm{IoU}_{ij}$ measures their 3D-box overlap. The child position is updated by:
$$
p_i^{t+1}=p_i^t+\vec{F}_{\mathrm{a}}+\sum_{j\neq i}\vec{F}_{\mathrm{r}}^{ij}.
$$
The implementation uses staged container expansion, boundary correction, overlap removal, clamping, and wall-facing orientation repair; details are provided in \cref{sec:fdlo_details}. During editing, the refinement is restricted to movable nodes whenever possible, preserving unrelated \mbox{scene regions intact}.

\subsection{Incremental Editing via LLM Agents}

The hierarchy gives editing an explicit retrieval unit: an instruction usually refers to an object, a support surface, or a local functional group rather than the whole scene. We therefore propose Hierarchical Retrieval-Augmented Generation (HRAG), inspired by RAG~\cite{Lewis20RAG}, to retrieve the relevant HDSL context before calling the LLM. For each node, we concatenate its identifier, class name, description, and ancestor path as the retrieval feature; raw coordinates and sizes are not embedded, but are included in the retrieved DSL passed to the editor. A BGE-family encoder~\cite{chen-etal-2024-m3} embeds both node features and the edit instruction. HRAG combines embedding similarity, lexical matches, and intent-aware constraints: rearrangement edits keep a wider local context, while remove and replace edits use a narrower target set. For rearrangement, the movable-object phrase is parsed separately from the destination phrase, so an explicit destination container guides context without replacing the target object.

The retrieved partial DSL contains the target nodes, their ancestors, and sibling context, since the LLM must see nearby occupancy to edit without disrupting the parent container. The edit prompt requires one XML-parsable HDSL fragment with the same root identifier, preserves unrelated siblings and asset metadata, and forbids strategy text or markdown. The edited fragment is then merged back into the original scene.

We formalize the editing algorithm with three components: semantic retrieval, localized rewriting, and subtree merging. Given a scene $\Phi(\mathcal{S}) = (\mathcal{N}, \mathcal{E})$ and parent function $\pi(n)=p$ when $(p,n)\in\mathcal{E}$, HRAG computes the relevant node set $\mathcal{N}_{\mathrm{rel}}$ for edit instruction $P_{\mathrm{edit}}$. Let $e_i=\mbox{Embed}(f(n_i))$ and $e_P=\mbox{Embed}(P_{\mathrm{edit}})$:

$$
\begin{aligned}
s_i &= 100 \cdot \frac{\langle e_i, e_P \rangle}{\|e_i\|\|e_P\|},\\
s_{\max} &= \max_{n_i \in \mathcal{N}} s_i ,
\end{aligned}
$$
where $f(n_i)$ denotes the path-aware textual feature of node $n_i$. Let $\delta_e$ be the intent-dependent similarity gap for $P_{\mathrm{edit}}$, and let $b_i$ be the keyword bonus from lexical overlap with the node identifier, class, and description. HRAG selects:
$$
\begin{aligned}
C_e &= \left\{ n_i \in \mathcal{N} \mid q_i \right\},\\
q_i &\equiv s_{\max} - s_i \leq \delta_e \ \lor\ b_i > 0,\\
\mathcal{N}_{\mathrm{rel}} &=
\operatorname{TopK}_{K_{\max}}\left(C_e, s_i+b_i\right).
\end{aligned}
$$
If fewer than $K_{\min}$ nodes are selected, we fall back to the top-$K_{\min}$ nodes by embedding similarity. We then construct the subtree $\Phi_{\mathrm{rag}}(\mathcal{S}) = (\mathcal{N}_{\mathrm{rag}}, \mathcal{E}_{\mathrm{rag}})$ that satisfies:
\begin{enumerate}
    \item $\mathcal{N}_{\mathrm{rel}} \subseteq \mathcal{N}_{\mathrm{rag}}$,
    \item $\mathcal{N}_{\mathrm{sib}} = \{ m \in \mathcal{N} \mid \exists n \in \mathcal{N}_{\mathrm{rel}}, \pi(n) = \pi(m) \} \subseteq \mathcal{N}_{\mathrm{rag}}$,
    \item $\forall n \in \mathcal{N}_{\mathrm{rag}}, \pi(n) \in \mathcal{N}_{\mathrm{rag}}$ if $\pi(n)$ exists.
\end{enumerate}

\begin{figure}[t]
    \centering
    \includegraphics[width=\columnwidth]{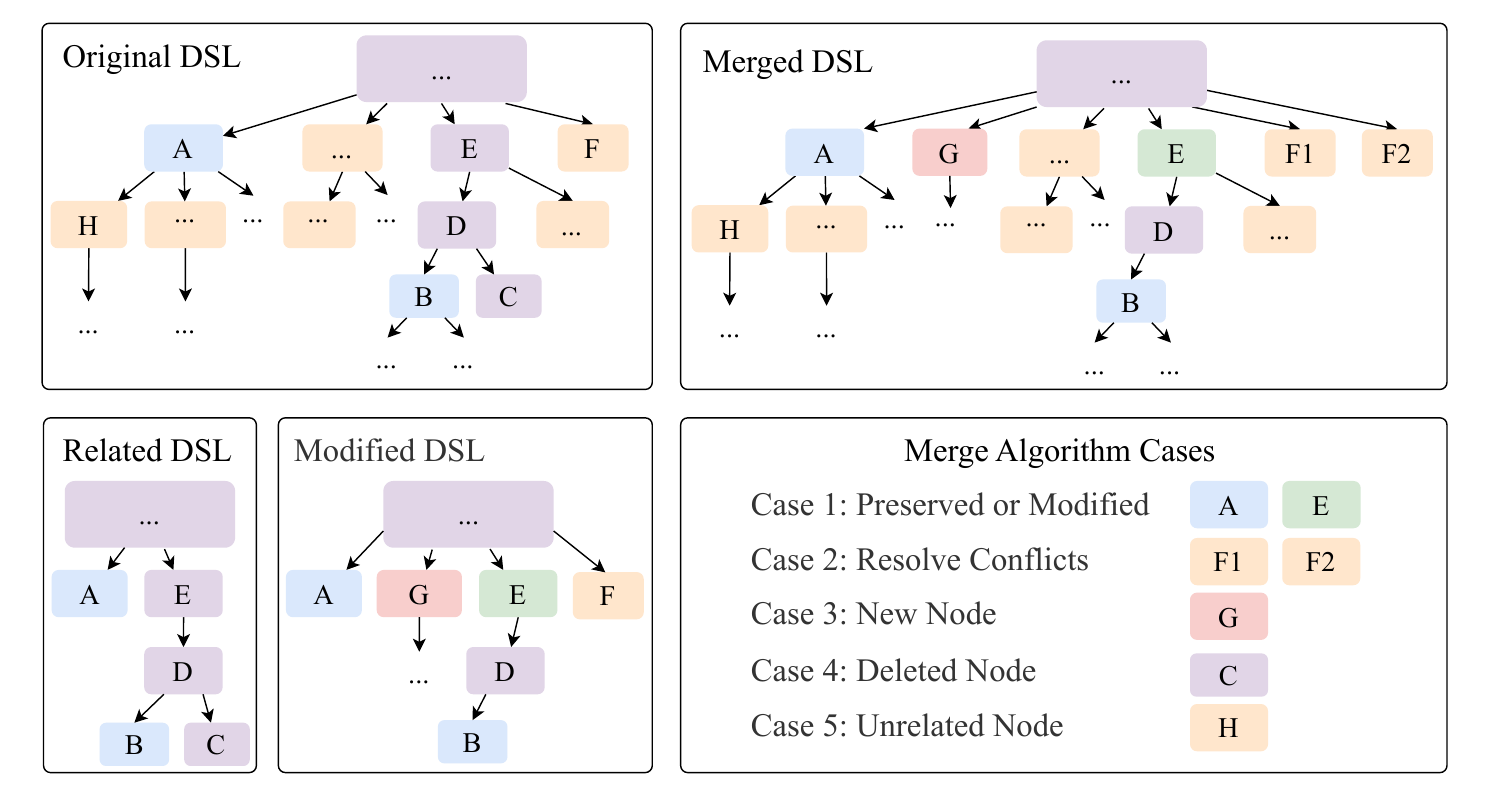}
    \caption{\textbf{Three-way merge.} The merge reconciles the original DSL, retrieved context, and edited subtree into a final HDSL while preserving unmodified nodes. The full algorithm is in the appendix.}
    \label{fig:merge}
\end{figure}

Next, the LLM edits $\Phi_{\mathrm{rag}}(\mathcal{S})$ to produce $\Phi_{\mathrm{rag}}'(\mathcal{S})$. Finally, we merge the edited subtree and apply the refinement pipeline to generate the updated scene. The merge algorithm performs a three-way reconciliation over the original $\Phi(\mathcal{S})$, the unedited partial subtree $\Phi_{\mathrm{rag}}(\mathcal{S})$, and the edited subtree $\Phi_{\mathrm{rag}}'(\mathcal{S})$, as shown in~\cref{fig:merge}. It applies edited attributes only to selected target subtrees, preserves original nodes included only as spatial context, allows deletion only for remove/replace intents, and inherits asset metadata for unchanged objects. The merge operation is:

$$
\mathrm{Merge}\left(\Phi, \Phi_{\mathrm{rag}}, \Phi_{\mathrm{rag}}'\right) \rightarrow \Phi_{\mathrm{final}}.
$$

Here, $\Phi_{\mathrm{final}}$ denotes the merged output, which is further refined in subsequent steps. The synthesized DSL, along with the scene configuration file, is then rendered using the simulation framework introduced by AI2-THOR~\cite{ai2thor}.

\section{Experiments}
\begin{table*}[t]
\caption{\textbf{Quantitative results on generation} across distinct scene categories.
The full-system comparison is between HDSL, Holodeck, and I-Design. Layout-only rows are marked with $\dagger$ and are geometry diagnostics, not entries in the full text-to-scene leaderboard. \textbf{Bold} and \textit{italics} mark best and second-best results among full scene-generation systems only.}
\centering
\resizebox{\textwidth}{!}{
\begin{tabular}{@{}ll ccccccc@{}}
\hline
Method & Metric & Apartment & Bedroom & Classroom & Gym & Office & Warehouse & \textbf{Avg.} \\
\hline
Holodeck~\cite{Yang_2024_CVPR_Holodeck} &  & \textbf{41.40} & \textbf{26.40} & \textit{36.71} & \textit{30.20} & \textit{77.60} & \textit{46.60} & \textit{43.15} \\
I-Design~\cite{celen2024idesign} & NObj$\uparrow$ & 9.00 & \textit{11.00} & 18.00 & 10.00 & 17.00 & 14.20 & 13.20  \\
DirectLayout\textsuperscript{\dag}~\cite{ran2025direct} &  & 5.60 & 5.00 & 27.60 & 8.40 & 9.40 & 6.20 & 10.37 \\
LayoutVLM\textsuperscript{\dag}~\cite{sun2025layoutvlm} &  & 13.00 & 8.00 & 28.00 & 14.00 & 25.00 & 22.00 & 18.33 \\
\textbf{HDSL (ours)} &  & \textit{25.67} & 7.43 & \textbf{54.80} & \textbf{52.75} & \textbf{135.08} & \textbf{73.40} & \textbf{58.19} \\
\hline
Holodeck~\cite{Yang_2024_CVPR_Holodeck}&  & \textit{0.13} & \textbf{0.05} & \textbf{0.00} & \textit{0.02} & \textit{0.15} & \textbf{0.00} & \textit{0.06}\\
I-Design~\cite{celen2024idesign} & OOB$\downarrow$ & 0.67 & 0.58 & 0.39 & 0.62 & 0.46 & 0.54 & 0.54  \\
DirectLayout\textsuperscript{\dag}~\cite{ran2025direct} &  & 0.00 & 0.04 & 0.00 & 0.12 & 0.00 & 0.00 & 0.03 \\
LayoutVLM\textsuperscript{\dag}~\cite{sun2025layoutvlm} &  & 0.00 & 0.12 & 0.00 & 0.21 & 0.12 & 0.05 & 0.08 \\
\textbf{HDSL (ours)} &  & \textbf{0.03} & \textit{0.15} & \textit{0.01} & \textbf{0.01} & \textbf{0.04} & \textit{0.07} & \textbf{0.05} \\
\hline
Holodeck~\cite{Yang_2024_CVPR_Holodeck} &  & 2.18 & 6.96 & 1.31 & 0.74 & 1.02 & 2.48 & 2.45\\
I-Design~\cite{celen2024idesign} & PIoU$\downarrow$ & \textbf{0.02} & \textbf{0.01} & \textbf{0.01} & \textbf{0.01} & \textbf{0.00} & \textbf{0.00} & \textbf{0.01}\\
DirectLayout\textsuperscript{\dag}~\cite{ran2025direct} &  & 0.75 & 0.04 & 0.43 & 0.37 & 1.68 & 0.28 & 0.59 \\
LayoutVLM\textsuperscript{\dag}~\cite{sun2025layoutvlm} &  & 0.61 & 0.09 & 0.03 & 0.98 & 0.03 & 1.40 & 0.52 \\
\textbf{HDSL (ours)} &  & \textbf{0.02} & \textit{0.03} & \textbf{0.01} & \textit{0.03} & \textit{0.01} & \textit{0.35} & \textit{0.07}  \\
\hline
Holodeck~\cite{Yang_2024_CVPR_Holodeck} &  & \textbf{23.53} & \textbf{26.05} & 17.26 & \textit{19.42} & 9.23 & \textbf{23.34} & \textit{19.81}\\
I-Design~\cite{celen2024idesign} & CLIP$\uparrow$ & 14.68 & \textit{25.47} & \textbf{26.03} & 13.34 & \textbf{15.93} & 22.03 & 19.58\\
\textbf{HDSL (ours)} &  & \textit{22.65} & 23.78 & \textit{24.55} & \textbf{22.12} & \textit{10.31} & \textit{22.51} & \textbf{20.99} \\
\hline
\hline
Holodeck~\cite{Yang_2024_CVPR_Holodeck} &  & 637.90 & 389.40 & 463.44 & 416.80 & 839.43 & 567.10 & 552.34 \\
I-Design~\cite{celen2024idesign} & Time$\downarrow$ &842.90  &687.10  &  871.80   & 925.20  &  813.60  &  952.70 & 848.90 \\
\textbf{HDSL (ours)} &   & \textbf{303.67} & \textbf{278.20} & \textbf{434.33} & \textbf{264.75} & \textbf{597.00} & \textbf{421.89} & \textbf{383.31}  \\
\hline
\end{tabular}
}
\vspace{0.5mm}
\begin{minipage}{0.98\textwidth}
\footnotesize
$\dagger$ DirectLayout and LayoutVLM are layout-only reproductions: NObj counts converted layout slots, and CLIP/time are omitted because these public pipelines do not include the same asset retrieval, rendering, or refinement stack.
\end{minipage}
\label{tab:benchmark}
\end{table*}

We evaluate \textbf{HDSL} through three questions: whether the representation improves text-prompted scene generation, whether hierarchical planning and FDLO account for the gains, and whether HRAG makes instruction-based edits cheaper and more local than rewriting the whole scene. Unless otherwise specified, HDSL uses Qwen2.5-72B-Instruct~\cite{qwen2025qwen25technicalreport}. The editing experiments are meant to test a model-interface property: whether an instruction can be grounded to the relevant structured context without forcing the LLM to rewrite unrelated scene content.

\subsection{Text-Prompted Generation}

\noindent\textbf{Baselines and scope.} We separate full text-to-scene generation from layout-only diagnostics. HDSL, Holodeck~\cite{Yang_2024_CVPR_Holodeck}, and I-Design~\cite{celen2024idesign} are complete systems with asset retrieval and rendering. DirectLayout~\cite{ran2025direct} and LayoutVLM~\cite{sun2025layoutvlm} are reported only as geometry diagnostics after conversion to our scene JSON format, so their NObj counts are layout slots rather than rendered objects. We select six MIT Scene Dataset categories~\cite{quattoni2009recognizing}; HDSL, Holodeck, I-Design, and DirectLayout use five runs per category, while LayoutVLM contributes one adapted layout per category. Holodeck room-size and I-Design termination guards are documented in the appendix. HSM~\cite{pun2026hsm} is excluded from the six-category main table because the reproducible public setup depends on HSSD-HAB support annotations and paired prompts; its three-prompt check is reported in the appendix.

\vspace{1mm}\noindent \textbf{Metrics.} Following~\cite{celen2024idesign, zhang2024SceneLanguage, tang2024diffuscene}, we evaluate generation with NObj, OOB (Out-of-Bounds), PIoU, CLIP, and generation time; formal definitions and timing scope are provided in the appendix.

\begin{figure*}[t]
    \centering
    \includegraphics[width=0.88\textwidth]{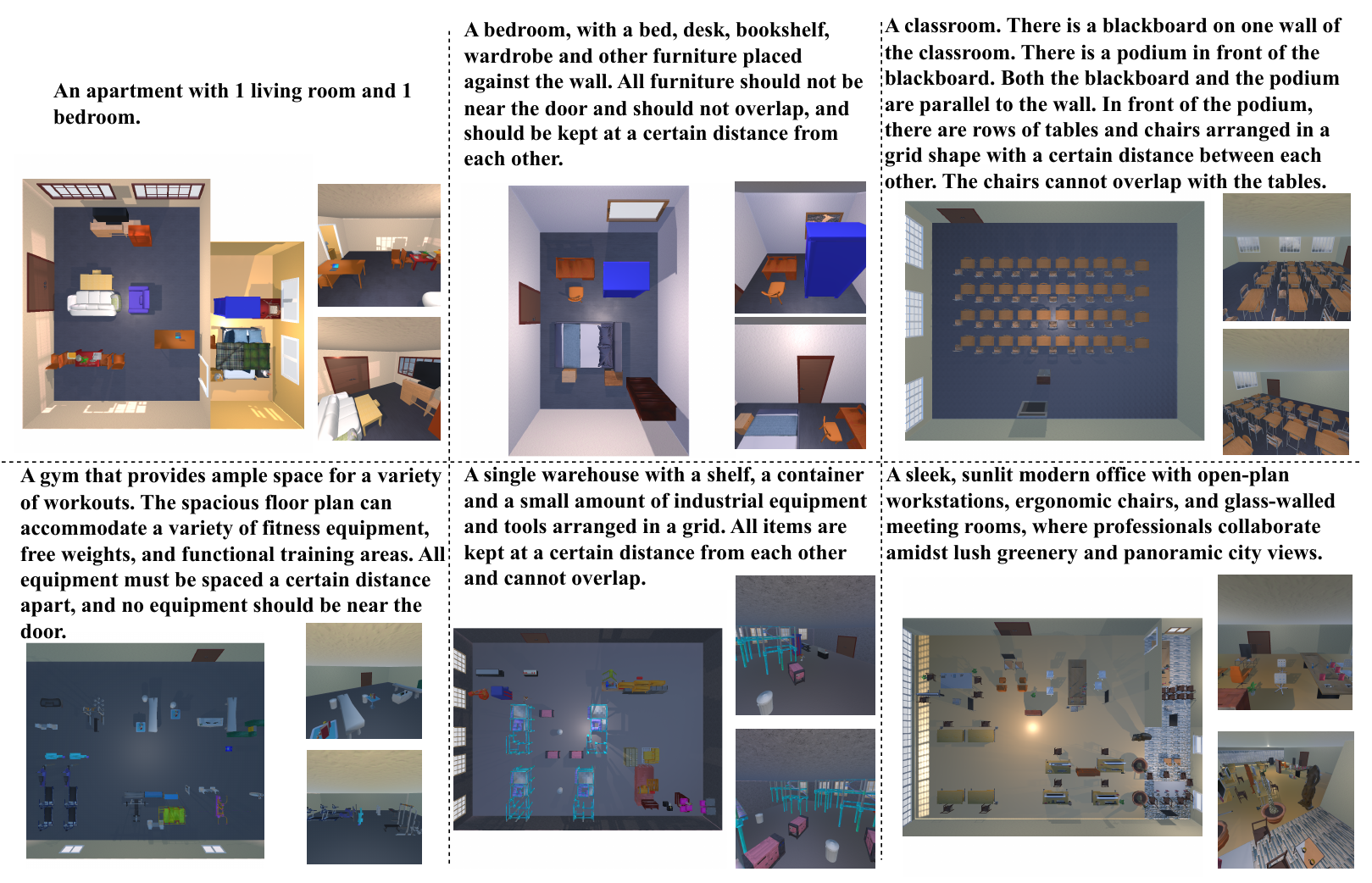}
    \caption{\textbf{Qualitative results on generation} of residential homes, commercial spaces, industrial settings, and public areas. The examples illustrate denser object coverage and more even use of large open areas.}
    \label{fig:comparison}
\end{figure*}

\vspace{1mm}\noindent \textbf{Results.} \cref{tab:benchmark} shows the main trade-off. Among full text-to-scene systems, HDSL has the highest average object count (58.19 versus 43.15 for Holodeck and 13.20 for I-Design), the best average CLIP score (20.99), and the lowest average generation time (383.31s). The gain is largest in offices, gyms, classrooms, and warehouses, where flat global planning often under-fills the room or places objects near walls. HDSL does not obtain the lowest PIoU overall because I-Design produces much sparser scenes, but it keeps PIoU low while adding far more objects. DirectLayout and LayoutVLM are not used to support end-to-end claims; they show that HDSL remains competitive on converted layout geometry while solving the harder asset-grounded scene-generation task. We further report HSM~\cite{pun2026hsm} in the appendix as a paired reproducibility check, including the partial public-pipeline case.

\subsection{Text-Prompted Editing}

\begin{figure*}[t]
    \centering
    \includegraphics[width=0.86\textwidth]{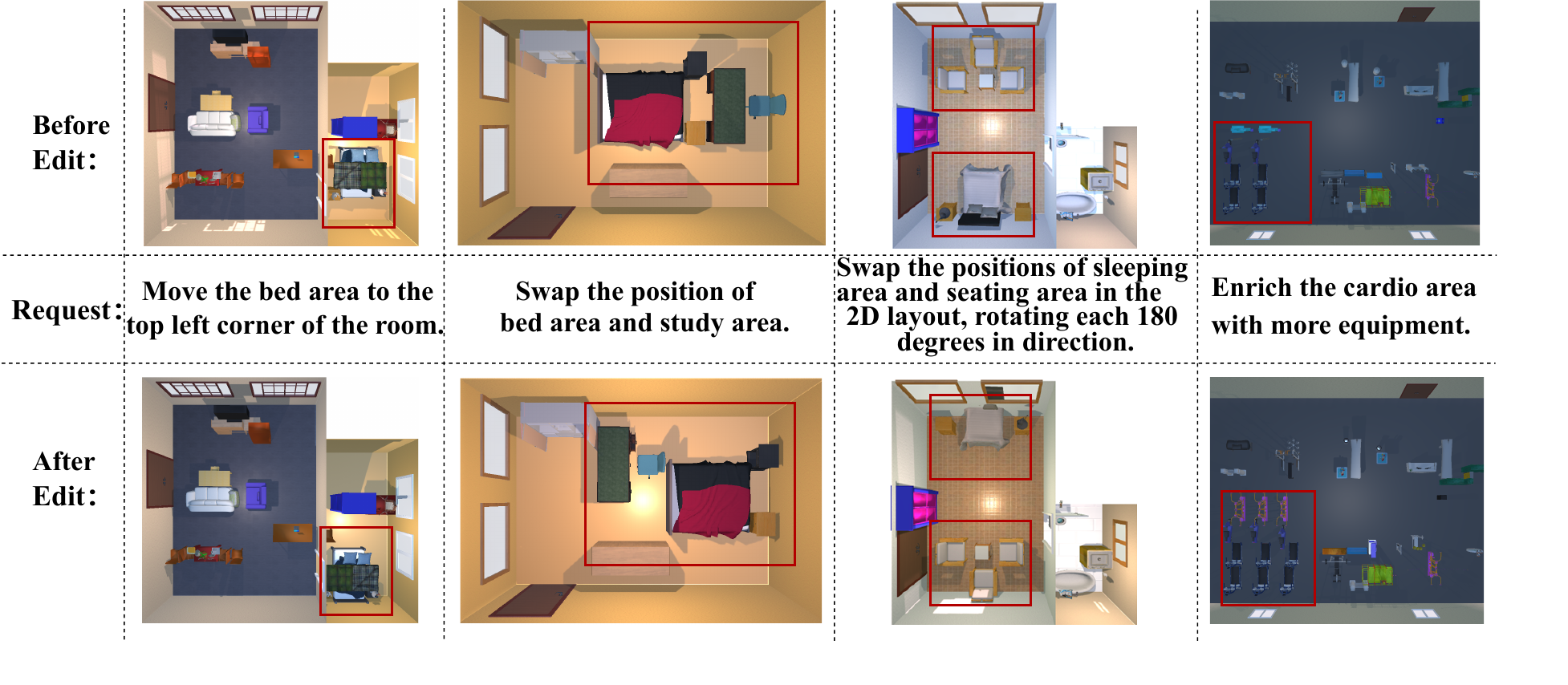}
    \caption{\textbf{Editing examples.} HDSL supports object movement, local swaps, scene enrichment, and reasoning-based edits while preserving nearby spatial relationships.}
    \label{fig:edit}
\end{figure*}

HDSL exposes an editable node for each object, support surface, and functional region, so natural-language instructions can be converted into local subtree rewrites rather than full-scene regeneration. \cref{fig:edit} shows object movement, local swaps, scene enrichment, and reasoning-based edits. The quantitative ablation below tests whether this locality preserves unrelated content.

\subsection{Ablation Study}

We perform ablation studies to test whether the gains come from the proposed representation and localization mechanisms. The experiments focus on hierarchical generation with refinement and partial editing through HRAG.

\vspace{1mm}\noindent \textbf{Ablation on Generation Components}. We evaluate three configurations on the same six scene types: the full model, direct HDSL generation without hierarchical generation (HG), and hierarchical generation without force-directed layout optimization (FDLO). Since LLM generation is stochastic, we compare relative trends within this controlled run. \cref{tab:ablation_gen} shows that HG increases object coverage and CLIP alignment, while FDLO mainly reduces boundary and overlap errors. Their combination gives the best overall trade-off.

\begin{table}[!htbp]
\centering
\caption{\textbf{Generation ablation} on pipeline components. HG increases object coverage and CLIP alignment, while FDLO reduces spatial violations.}
\label{tab:ablation_gen}
\footnotesize
\setlength{\tabcolsep}{4.0pt}
\renewcommand{\arraystretch}{0.96}
\resizebox{\columnwidth}{!}{
\begin{tabular}{lcccc}
\hline
HG   & $\times$ & $\surd$ & $\times$ & $\surd$ \\
FDLO & $\times$ & $\times$ & $\surd$  & $\surd$ \\
\hline
NObj $\uparrow$  & 8.46            & \textit{55.82}  & 14.10           & \textbf{65.23} \\
OOB $\downarrow$ & 0.40            & 0.36            & \textbf{0.03}   & \textit{0.10}  \\
PIoU $\downarrow$& \textbf{0.03}   & \textbf{0.03}   & 0.09            & \textit{0.05}  \\
CLIP $\uparrow$  & 17.62           & \textit{20.71}  & 16.96           & \textbf{20.76} \\
\hline
\end{tabular}
}
\end{table}

\vspace{1mm}\noindent \textbf{Ablation on Editing Mechanism}.
We compare HRAG-HDSL with a full-HDSL rewrite baseline that gives the LLM the whole scene and asks for a complete edited scene. The eight paired edits use the same layouts and instructions for both methods. As shown in~\cref{tab:edit_repr_ablation}, geometric validity alone is not enough: a rewritten scene can satisfy OOB and PIoU checks while deleting objects or moving unrelated furniture. Full-scene rewriting uses $5.22\times$ more tokens and $6.19\times$ more time, falls back in $50.0\%$ of cases, removes $1.50$ objects on average, and moves $23.8\%$ of stable non-target objects. HRAG-HDSL produces valid DSL in every case, uses no fallback, removes no objects, and keeps stable non-target displacement at $0.018$m. Because a fixed $0.15$m threshold is size-sensitive, we also report continuous target displacement: HRAG moves targets by $0.942$m on average versus $0.290$m for full rewriting. The three HRAG threshold misses remain valid, target-preserving edits with small post-refinement displacement; the user study below evaluates semantic instruction compliance directly.

\begin{table}[!htbp]
\centering
\caption{\textbf{Quantitative ablation of HRAG for instruction-based editing} on eight paired edits. Target changed uses a strict $0.15$m displacement threshold, while Target $\Delta$ reports continuous target motion; preservation metrics exclude the target and post-processing repair steps used by refinement.}
\small
\setlength{\tabcolsep}{4.0pt}
\renewcommand{\arraystretch}{0.98}
\begin{tabular}{@{}lcc@{}}
\toprule
\textbf{Metric} & \textbf{HRAG-HDSL} & \textbf{Full HDSL rewrite} \\
\midrule
Target changed $\uparrow$ & \textbf{62.5\%} & 50.0\% \\
Target $\Delta$ $\uparrow$ & \textbf{0.942m} & 0.290m \\
Valid DSL $\uparrow$ & \textbf{100.0\%} & 50.0\% \\
Fallback $\downarrow$ & \textbf{0.0\%} & 50.0\% \\
Tokens $\downarrow$ & \textbf{4.1k} & 21.4k \\
Runtime $\downarrow$ & \textbf{72.1s} & 445.8s \\
Stable $\Delta$ $\downarrow$ & \textbf{0.018m} & 0.200m \\
Stable moved $\downarrow$ & \textbf{4.3\%} & 23.8\% \\
Objects removed $\downarrow$ & \textbf{0.00} & 1.50 \\
\bottomrule
\end{tabular}
\label{tab:edit_repr_ablation}
\end{table}

\begin{figure}[!htbp]
\centering
\includegraphics[width=\columnwidth]{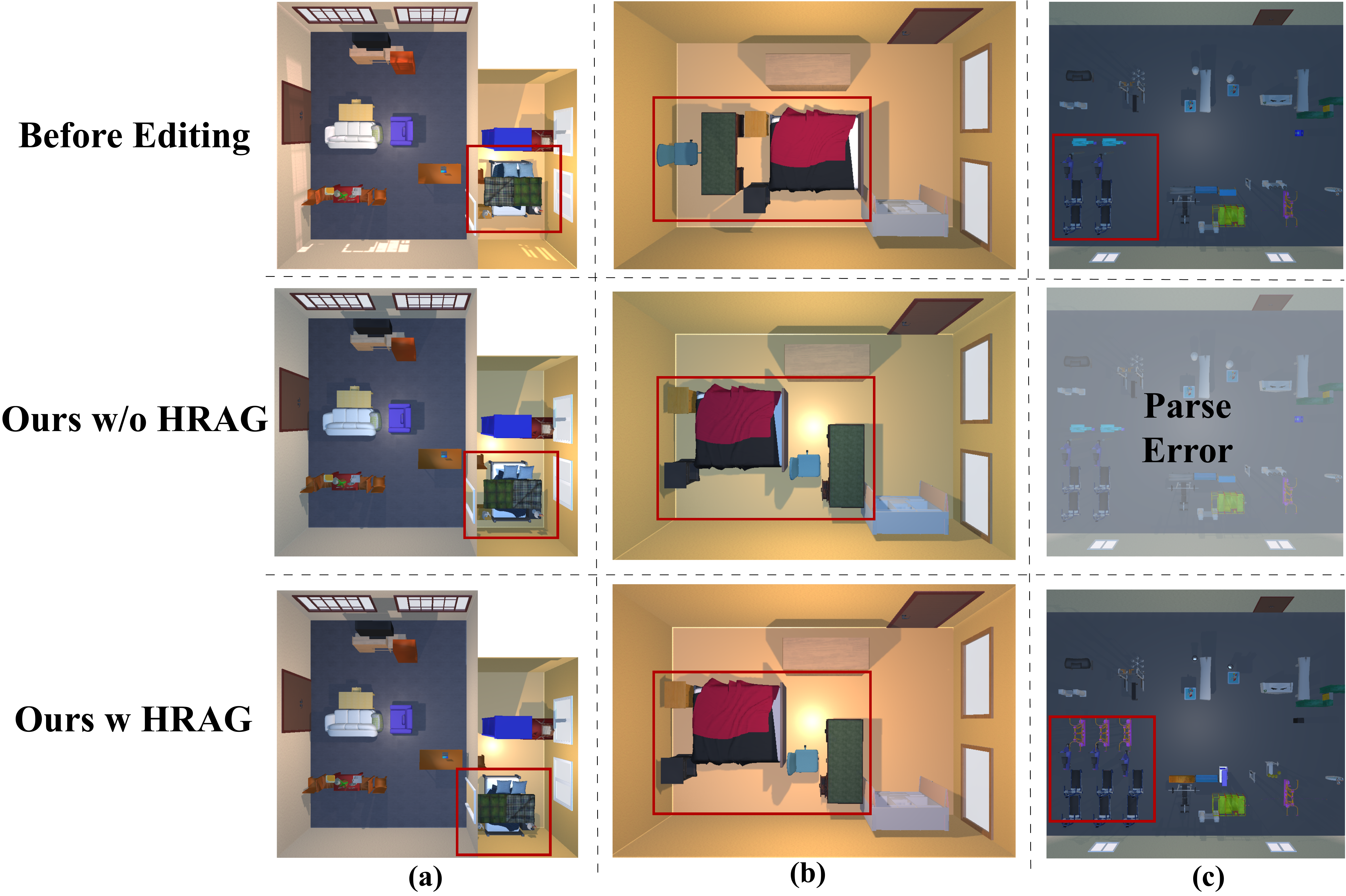}
\caption{\textbf{Editing ablation examples.} Full-scene rewriting can miss a local edit, work in simple scenes, or become unstable in complex scenes with long context, while HRAG edits the retrieved local subtree.}
\label{fig:edit_ablation}
\end{figure}

\subsection{User Study}
\noindent We recruited 20 participants for a perceptual study. Participants were general users rather than method authors, did not see system names, and were not required to have specialized 3D modeling background. For generation, each participant evaluated six anonymized triplets from HDSL, Holodeck~\cite{Yang_2024_CVPR_Holodeck}, and I-Design~\cite{celen2024idesign}, giving 120 votes that we convert into Bradley--Terry scores~\cite{19ff28b9-64f9-3656-ba40-08326a05748e}. For editing, participants rated nine rendered edits on a three-point compliance scale, giving 180 votes; $82.2\%$ judged the edit as exactly or mostly matching the instruction, which complements the automatic displacement and preservation metrics.

\begin{figure}[!htbp]
\centering
\includegraphics[width=\columnwidth]{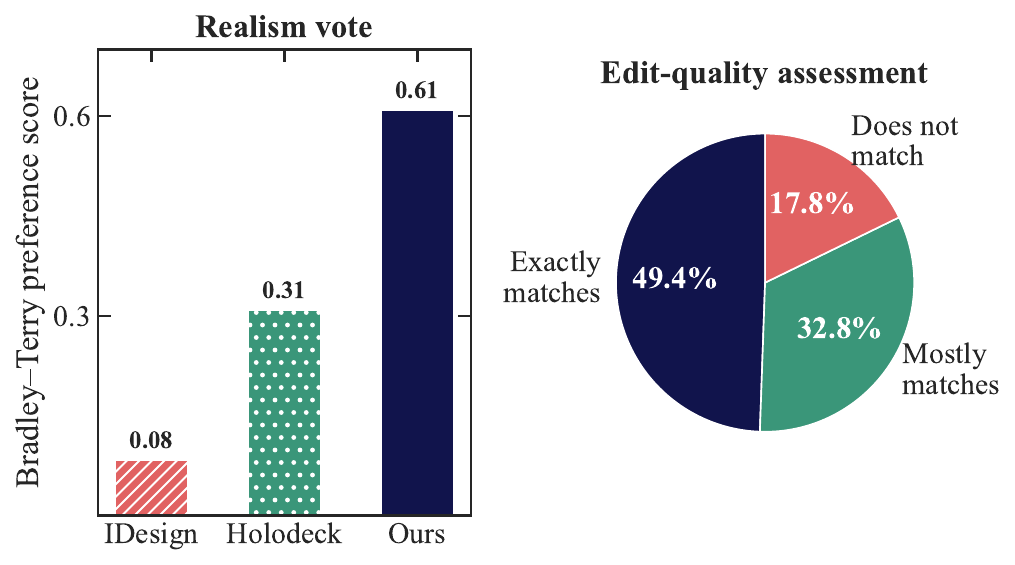}
\caption{\textbf{User study results.} Participants prefer HDSL generations over Holodeck and I-Design, and $82.2\%$ of editing votes rate the result as exactly or mostly matching the editing instruction overall.}
\label{fig:user_study}
\end{figure}

\FloatBarrier

\section{Conclusion}

We presented HDSL, a hierarchical DSL that turns 3D indoor scene generation and editing into bounded structured prediction for LLM agents. The same representation supports recursive generation, deterministic checking, asset grounding, force-directed refinement, and HRAG-based local editing. Across generation, ablation, and editing experiments, HDSL improves scene coverage and edit efficiency while preserving a readable node-level interface for inspection and repair.

\clearpage
\section*{Limitations}
This work focuses on language-mediated scene structure: how an LLM can write, check, retrieve, and locally revise a 3D indoor scene. HDSL assumes a 3D asset database with reasonable category coverage, so rare-object visual fidelity depends on the retrieval corpus even when geometry is valid. FDLO targets static indoor layouts by repairing bounding-box collisions and boundary violations; dynamics, articulation, and task-level affordances can be layered through simulators or specialized planners. HRAG is most direct for edits with identifiable spatial targets, while broad style changes may retrieve larger contexts or use full-scene rewriting. Recent baselines are reported under public-release boundaries, and editing success is measured with strict displacement diagnostics alongside human compliance judgments for edits.

\bibliography{main}

\clearpage

\appendix

\begin{center}
{\Large\bfseries Appendix}
\end{center}
\vspace{0.5em}
\normalsize\normalfont
\section{Additional Implementation Details}

\subsection{Floorplan-to-DSL Conditioning}
Throughout the generation process, we first follow the Holodeck methodology to generate floorplans based on user-provided prompts, including the positions of rooms, windows, and doors. This information is subsequently converted into the description field within the DSL attributes for later hierarchical scene generation and editing.

\subsection{From 2D CSS Layout to 3D Poses}
\label{sec:hdsl_lifting}
Each node uses a 2D CSS footprint in the parent plane and is lifted to 3D by fitting the retrieved asset's 3D bounding box to the corresponding floor region with rotation. The CSS coordinates define the $(x,z)$ ground-plane projection, while the vertical coordinate $y$ is determined through \textbf{place-on semantics}: each child lies on the supporting surface of its parent, with $y = y_{\mathrm{parent}} + h_{\mathrm{surface}}$. For objects with multiple supporting surfaces, such as bookshelves, the parent is expanded into child virtual containers whose geometry provides the corresponding vertical offset. The CSS \texttt{transform: rotate} property maps to rotation around the vertical axis, and the retrieved asset is scaled to the node's \texttt{size} attribute.

\subsection{HDSL Node Schema and Example}
\begin{table}[!htbp]
\caption{\textbf{Specification} of HDSL node attributes.}
\centering
\small
\resizebox{\columnwidth}{!}{
\begin{tabular}{@{}ll@{\hspace{1.5em}}ll@{}}
    \toprule
    \textbf{Attr.} & \textbf{Description} & \textbf{Attr.} & \textbf{Description} \\
    \midrule
    \texttt{id} & A unique identifier. & \texttt{virtual} & Whether it binds to a retrieved asset. \\
    \texttt{style} & CSS properties specifying the layout. & \texttt{is\_leaf} & Whether it is a leaf. \\
    \texttt{class} & Specifies the asset class name. & \texttt{variance\_type} & Variance type of sibling assets. \\
    \texttt{size} & The 3D bounding-box size. & \texttt{description} & A textual description of the node.\\
    \bottomrule
    \end{tabular}
}
\label{tab:hdslasspec}
\end{table}

HDSL follows an HTML/CSS-like structure: a scene is composed of nested div-style elements, and each element represents a room, region, object, or virtual support container. This choice uses LLMs' prior exposure to web code while keeping coordinates local to each parent. It also matches indoor-scene structure, where rooms decompose into zones, zones decompose into object groups, and gravity creates many place-on relations.

\subsection{LLM Setup and Sanity Checks}
\label{sec:llm_sanity}
We employ Qwen2.5-72B-Instruct~\cite{qwen2025qwen25technicalreport} as the base model for the reported LLM-agent scene generation and editing runs. The executable generation prompt requires the model to return exactly one XML-parsable HDSL fragment rooted at the requested node, with no natural-language strategy, markdown fence, heading, or explanatory text. We adopt a bounded multi-candidate generate--verify--revise loop for quality control. At each node, the implementation samples multiple candidates, parses them into HDSL, applies deterministic repairs when safe, and then scores candidate quality. Regeneration is triggered by malformed XML, duplicate identifiers, root mismatch, children outside parent bounds, invalid support nesting, emitted clearance/aisle nodes, excessive overlap, door/window blockage, implausible orientation, unrealistic asset-size assumptions, missing functional coverage, or extreme occupancy. Each node has a maximum attempt budget; if the budget is exceeded, the best valid candidate is used when available, otherwise the current branch is terminated for safety.

\subsection{Parallel Expansion Policy}
Parallelization is limited to recursive expansion of disjoint subtrees, not sibling generation. All siblings are generated jointly in a single LLM call, enabling explicit spatial coordination and avoiding intra-container conflicts. Subsequent parallel expansion is safe since each branch remains within its parent's footprint. A final force-directed layout refinement resolves residual overlaps and boundary violations during recursive refinement.

\subsection{Force-Directed Layout Optimization Details}
\label{sec:fdlo_details}
FDLO runs independently inside each parent node and is then applied recursively down the hierarchy. The attraction term uses $\vec{D}_i$ to move an out-of-bound child toward the valid interior of its parent. The repulsion term uses $\vec{\Delta}_{ij}$ to separate overlapping siblings, scaled by their 3D bounding-box $\mathrm{IoU}_{ij}$ so larger collisions receive stronger corrections. In practice, we first expand virtual containers when their children exceed the current footprint, then run boundary-focused refinement, balanced boundary-overlap refinement, and a final fine pass. The optimizer uses adaptive force scaling, boundary clamping, overlap post-processing, and automatic orientation correction for wall-adjacent objects. During editing, the same refinement pipeline is constrained to the movable subset when possible, which keeps unrelated scene~regions stable. The main ablation quantifies the effect: adding FDLO to direct HDSL generation reduces OOB from $0.40$ to $0.03$, while adding FDLO to hierarchical generation reduces OOB from $0.36$ to $0.10$ and keeps PIoU low ($0.03$ to $0.05$) despite denser scenes.

\subsection{Implementation hyperparameters}

The hyperparameters are as follows:

\begin{tcolorbox}[mysinglecolumnbox]
generation\_max\_attempts = 5 \\
generation\_num\_candidates = 2 \\
generation\_accept\_score = 0.58 \\
occupancy\_ratio\_virtual\_min = 0.12 \\
occupancy\_ratio\_preferred\_max = 0.8 \\
occupancy\_ratio\_hard\_max = 0.9 \\
min\_expand\_area = 900 \\
max\_expand\_depth = 5 \\
retrieve\_top\_k = 10 \\
related\_editing\_threshold = 5 \\
edit\_min\_selected\_nodes = 2 \\
edit\_max\_selected\_nodes = 8 \\
edit\_max\_attempts = 6 \\
FDLO\_iterations = 500 \\
FDLO\_base\_repulsion\_force = 0.02 \\
FDLO\_boundary\_force\_multiplier = 10.0 \\
FDLO\_overlap\_iou\_eps = 0.02 \\
FDLO\_overlap\_post\_process\_passes = 8 \\
FDLO\_virtual\_container\_fit\_margin = 0.20 \\

\end{tcolorbox}

\noindent\textbf{Candidate Quality and Occupancy}. For each generated node, the implementation computes child-area occupancy and uses it as one component of a broader quality score. For virtual containers, occupancy below $0.12$ is rejected as under-specified; occupancy up to $0.8$ is considered preferred, and occupancy above $0.9$ is rejected as overcrowded. For concrete non-leaf objects, the occupancy check is relaxed because small supported items may occupy only a small portion of the support surface. The final acceptance decision also considers bounds validity, sibling overlap, distribution, support semantics, door/window clearance, wall-facing orientation, functional coverage, and asset-size realism.

\noindent\textbf{Expansion Check Threshold}. Nodes with an area smaller than $min\_expand\_area$ are treated as leaves. Furthermore, if the current node's depth exceeds $max\_expand\_depth$, expansion is halted to maintain computational efficiency and prevent over-segmentation of the scene hierarchy. Structural storage nodes such as shelves and bookcases may also be kept as leaves to avoid generating unsupported internal artifacts.

\noindent\textbf{Object Retrieval Threshold}. We use Objaverse~\cite{deitke2023objaverse} assets and follow the Holodeck retrieval pipeline~\cite{Yang_2024_CVPR_Holodeck}. The retriever combines OpenCLIP~\cite{openclip} visual-text similarity, Sentence-BERT~\cite{reimers-2019-sentence-bert} textual similarity, category constraints, physical-size checks, and support-surface compatibility. Candidate retrieval is thresholded by semantic similarity and falls back to catalog-level category matches when strict retrieval is empty. The candidate set is capped by $retrieve\_top\_k=10$ before asset assignment.

\noindent\textbf{Related Editing Node Threshold}. The released implementation uses a BGE-small-en-v1.5 encoder by default for HRAG node retrieval. For each node, the retrieval text concatenates its identifier, class, description, and ancestor context; coordinates and sizes are not part of the embedding text, but remain available in the retrieved HDSL fragment. Similarities are scaled by $100$, and all nodes whose score is within $related\_editing\_threshold=5$ of the maximum score are considered candidates. The gap is intent-aware: rearrangement uses a slightly wider context, while remove/replace uses a narrower one. For rearrangement, relation markers such as ``closer to" and ``to the center" are used to extract the movable-object phrase before container filtering, so a destination region does not override the object being moved. We match explicit node identifiers exactly when they appear in the movable phrase, add a lexical keyword bonus for direct matches with object ids/classes/descriptions, prune to direct target objects, and cap the final set to $edit\_max\_selected\_nodes=8$ with a fallback minimum of $edit\_min\_selected\_nodes=2$.

\noindent\textbf{FDLO Defaults}. FDLO is implemented as staged recursive refinement. Each parent node is optimized for up to $500$ iterations with base repulsion $0.02$, boundary force multiplier $10.0$, overlap IoU epsilon $0.02$, and $8$ overlap post-processing passes. Before iterative refinement, virtual containers can be expanded to fit their children with a $0.20$ margin and $1.25$ area slack. After refinement, the implementation applies boundary clamping and automatic orientation correction for wall-adjacent objects after layout optimization.

\subsection{Evaluation Metrics}
Following prior text-to-scene evaluation protocols, we report five metrics. NObj counts placed object instances for full text-to-scene systems and converted layout slots for layout-only diagnostics. OOB (Out-of-Bounds) is the fraction of generated objects intersecting room boundaries. PIoU measures inter-object collision as the average pairwise 3D bounding-box intersection-over-union. CLIP uses CLIP similarity~\cite{radford2021clip,hessel-etal-2021-clipscore} between rendered images and text prompts to estimate text-scene alignment. Time is wall-clock time from the start of scene generation to a saved renderable scene, including LLM calls, parsing, asset retrieval, layout refinement, and scene serialization; it excludes one-time downloads and offline metric computation.

For recent baselines whose public code does not produce the same AI2-THOR scene configuration, we convert saved layouts into a common scene JSON and run the same NObj, OOB, and PIoU metrics. We do not report CLIP or end-to-end runtime for layout-only reproductions unless the baseline includes comparable asset retrieval, rendering, and refinement stages. All full-system timing rows were produced in the same local experiment environment, while layout-only rows are used only for geometry diagnostics in the table.

\subsection{Editing Ablation Protocol}
The editing ablation compares HRAG-HDSL with a full-HDSL rewrite baseline under the same base scenes and edit instructions. HRAG-HDSL retrieves the relevant subtree, asks the LLM to rewrite only that local context, and merges the result back. The full-HDSL rewrite baseline gives the LLM the complete HDSL scene and asks for a full edited scene, using the same model, asset reuse setting, coordinate recalculation, and refinement pipeline.

We evaluate three groups of signals. Target changed measures whether the intended object moves by at least $0.15$m. This threshold is a strict furniture-scale diagnostic for the eight edit cases; it is not a universal semantic-success threshold and would be too coarse for small tabletop objects. We therefore also report continuous target displacement in the main table. Locality measures the mean displacement and moved rate of non-target objects; stable preservation excludes the target object and objects that were explicitly moved by post-processing repair. Metadata preservation checks whether unchanged objects keep their asset identifiers and retrieved asset metadata. This protocol is deliberately stricter than geometric validity alone: a method can have low OOB and PIoU while still failing the editing task by deleting objects or moving unrelated furniture items.

\paragraph{Expanded editing stress check.}
We evaluate eight paired edit instructions across apartments, classrooms, restaurant kitchens, and studio scenes. This run tests whether local editing remains parseable and content-preserving under the same layouts and instructions used for the full-HDSL rewrite baseline. For HRAG-HDSL, all cases produce valid DSL without fallback, preserve asset metadata, and remove no objects. Under the strict $0.15$m target-change threshold used in the main table, five of eight cases pass; the remaining cases are small-displacement edits where the target remains preserved but moves less than the threshold. As shown in Table~\ref{tab:expanded_hrag_edit}, local editing uses far fewer LLM tokens and much less time while reducing non-target displacement and object removal.

\begin{table}[!htbp]
\centering
\footnotesize
\setlength{\tabcolsep}{2.0pt}
\renewcommand{\arraystretch}{0.95}
\resizebox{\columnwidth}{!}{%
\begin{tabular}{@{}lccccccc@{}}
\toprule
\textbf{Method} & \textbf{Cases} & \textbf{Target} & \textbf{Valid/fallback} & \textbf{Stable $\Delta$} & \textbf{Moved} & \textbf{Removed} & \textbf{Cost} \\
\midrule
HRAG-HDSL & 8 & 62.5\% & 100.0\% / 0.0\% & 0.018m & 4.3\% & 0.00 & 4.1k / 72.1s \\
Full HDSL rewrite & 8 & 50.0\% & 50.0\% / 50.0\% & 0.200m & 23.8\% & 1.50 & 21.4k / 445.8s \\
\bottomrule
\end{tabular}
}
\caption{Expanded editing stress check on eight paired instructions. Target reports the fraction of cases whose intended object moves by at least $0.15$m; valid/fallback reports DSL parse success and fallback rate, and stable $\Delta$ and moved rate are computed on stable non-target objects after editing.}
\label{tab:expanded_hrag_edit}
\end{table}

\noindent\textbf{Stress-check pattern.}
The non-exact HRAG cases in this stress check are threshold misses rather than parser or object-deletion errors: the intended object remains present and metadata-preserved, but its displacement is smaller than $0.15$m after refinement. The three cases are the apartment coffee-table edit ($0.144$m), the classroom teacher-desk edit ($0.054$m), and the restaurant sink edit ($0.039$m). Full-scene rewriting shows a different profile. Four of eight full rewrites fall back to the original scene after the attempt budget, and two valid classroom rewrites remove six non-target objects each. Even when the output is geometrically valid, unrelated objects are more likely to move or disappear. This difference is why we report target change, parser reliability, and preservation metrics instead of relying only on OOB and PIoU.

\subsection{Reproducibility Scope of Recent Baselines}
DirectLayout and LayoutVLM are evaluated as geometry-only reproductions because their released pipelines do not expose the same complete text-to-render stack as Holodeck, I-Design, or HDSL. DirectLayout is run through its public prompt-to-layout branch. LayoutVLM is adapted with a fixed asset set per prompt and its released constraint-generation and optimization loop, but without Blender/image feedback. Their rows should therefore be read as layout diagnostics, not as full-scene baselines. For Holodeck, we only raise the maximum room-size limit so large MIT-style prompts are not rejected by a hard cap. For I-Design, we add a forced termination and recursion-depth guard so repeated layout-modification loops return the best available state instead of hanging; prompts, model, and layout rules are otherwise kept unchanged. HSM is evaluated separately on paired prompts because it depends on HSSD-HAB assets and support-surface annotations; the full run is marked when a prompt produces only a partial state. These settings make the comparisons reproducible while keeping the reported claims tied to the part of each baseline that can be run locally.

\subsection{Audit Trail for Reported Numbers}
All reported quantitative values are recomputed from saved scene JSON, JSONL edit records, and CSV summaries rather than from transient in-memory objects. Generation metrics are computed by the same JSON metric runner after each method's output is converted into the common scene format. The editing runs record the selected node ids, direct movable ids, expanded movable ids, valid-DSL flag, fallback flag, token count, runtime, and output scene path for each method-case pair. The locality and metadata tables are produced by separate post-processing scripts, which lets us recompute target displacement, non-target preservation, removed-object counts, and asset-metadata preservation after a run finishes. The experiment scripts skip completed cases on rerun and append missing records, so interrupted batches can be resumed without overwriting confirmed outputs. For recent layout-only baselines, we first convert released outputs into the common scene JSON format and then call the same metric code used for HDSL. These files define the scope of each comparison and prevent partial baseline reproductions from being reported as full text-to-render systems. We report averages over the available runs rather than inferential significance tests, since the expensive generation baselines provide limited repeated samples in public runs.

\subsection{Evaluation Boundaries}
CLIP is a proxy for text-scene alignment, so we pair it with human preference judgments and geometry metrics rather than treating it as a standalone score. The 20-participant user study is used as a rendered-scene consistency check for the automatic metrics; broader population-level preference studies are a natural next step once larger repeated generations are available. Baseline reproducibility is handled through explicit reporting boundaries: DirectLayout and LayoutVLM are evaluated on the layout portion supported by their public releases, while HSM is reported on paired prompts because its full pipeline depends on HSSD-HAB support annotations and VLM feedback. We therefore reserve the strongest claims for comparisons with full text-to-scene systems and use recent layout-only baselines as geometry stress tests. For editing, the $0.15$m target-change threshold is a strict diagnostic, so we report continuous target displacement, valid-DSL rate, fallback rate, non-target displacement, human compliance, and object removal as separate signals in the analysis.

\subsection{FDLO Refinement Stability}
\begin{figure}[!htbp]
    \centering
    \includegraphics[width=\columnwidth]{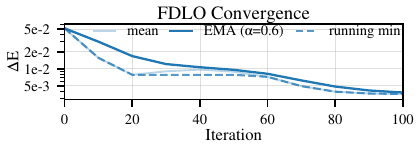}
    \caption{\textbf{FDLO convergence.} On a multi-room complex scene, the update magnitude decreases smoothly and plateaus without sustained oscillations.}
    \label{fig:fdlo_convergence_suppl}
\end{figure}
The plotted value measures the magnitude of each refinement update during iterative optimization. The monotonic decay and later plateau show \mbox{stable convergence behavior} rather than oscillation in challenging, large-scale layouts. The component ablation gives the corresponding final metrics: with direct generation, FDLO reduces OOB from $0.40$ to $0.03$; with hierarchical generation, it reduces OOB from $0.36$ to $0.10$ while preserving low PIoU ($0.05$ after refinement).

\subsection{Additional Model Comparison and Ablation}
We additionally evaluate recent LLMs under the same generation setting and report average OOB/PIoU and NObj/CLIP in Table~\ref{tab:model_comparison_suppl}.

\begin{table}[!htbp]
\centering
\footnotesize
\setlength{\tabcolsep}{2.0pt}
\renewcommand{\arraystretch}{0.9}
\resizebox{\columnwidth}{!}{%
\begin{tabular}{@{}lcccc@{}}
\toprule
\textbf{Metric} & \textbf{Q3} & \textbf{G3} & \textbf{GPT} & \textbf{DS} \\
\midrule
OOB / PIoU $\downarrow$
& 3.29 / 0.94
& 3.47 / 0.86
& \textbf{1.28} / \textbf{0.78}
& 2.47 / 1.04 \\
NObj / CLIP $\uparrow$
& 84.50 / \textbf{23.54}
& 67.50 / 22.81
& 65.33 / 20.97
& \textbf{102.17} / 22.99 \\
\bottomrule
\end{tabular}
}
\caption{Additional comparison with recent LLMs: Q3 denotes Qwen3-max, G3 denotes Gemini3-Pro, GPT denotes GPT-5.2, and DS denotes DeepSeek-v3.2.}
\label{tab:model_comparison_suppl}
\end{table}

\paragraph{Recent hierarchical baseline.}
We further compare HDSL with HSM~\cite{pun2026hsm}, a recent hierarchical scene-motif generator. We reproduce two HSM settings on three paired indoor prompts. The first uses the large-object branch, which is the most stable public setting. The second uses the full released pipeline after adding the official HSSD-HAB support-surface package, enabling large, wall, ceiling, and small objects. The full setting completes two prompts and produces a partial state for the home-office prompt after repeated invalid VLM responses during floor support placement; we include this partial state in the geometry evaluation and mark the row accordingly. This comparison is meant as a reproducibility check under matched prompts rather than a full benchmark. As shown in Table~\ref{tab:hsm_hdsl_repro}, HDSL matches the full HSM run in object count while producing fewer boundary violations and collisions with lower runtime.

\begin{table}[!htbp]
\centering
\footnotesize
\setlength{\tabcolsep}{2.4pt}
\renewcommand{\arraystretch}{0.95}
\resizebox{\columnwidth}{!}{%
\begin{tabular}{@{}lcccc@{}}
\toprule
\textbf{Method} & \textbf{NObj $\uparrow$} & \textbf{OOB $\downarrow$} & \textbf{PIoU $\downarrow$} & \textbf{Time $\downarrow$} \\
\midrule
HSM-large~\cite{pun2026hsm} & 6.00 & 0.1698 & 0.5223 & 309.0s \\
HSM-full~\cite{pun2026hsm}\textsuperscript{\ddag} & 11.67 & 0.3653 & 0.8708 & 654.5s \\
\textbf{HDSL (ours)} & \textbf{11.67} & \textbf{0.0000} & \textbf{0.0000} & \textbf{215.7s} \\
\bottomrule
\end{tabular}
}
\caption{Paired comparison with reproducible HSM settings on three indoor prompts. $\ddagger$ The full HSM row includes two completed cases and one partial home-office scene; runtime is averaged over the two completed cases in total.}
\label{tab:hsm_hdsl_repro}
\end{table}

\FloatBarrier
\section{Three-Way Merge Details}
Naive subtree replacement can fail because retrieved DSLs for editing are partial skeletons: to reduce context length, some contextual nodes may be included as leaves with their descendants pruned. Direct replacement would overwrite full nodes with these pruned versions, causing information loss. Our three-way merge shown in~\cref{alg:merge} preserves components present in the Original but missing in the Retrieved/Edited, while applying target edits and resolving identifier conflicts. Attribute conflicts are resolved by the selected target ids: edited attributes are accepted only for selected target subtrees, context-only nodes keep the original attributes, new edited children are accepted only for add/replace intents, and deletion is allowed only when a removed subtree contains a selected target under remove/replace intent. Unchanged objects keep their original retrieved asset metadata intact across all merge steps.

\begin{algorithm}[!htbp]
\caption{Guarded Three-Way Merge}
\label{alg:merge}
\begin{algorithmic}[1]
\REQUIRE Original tree $\Phi(\mathcal{S})$, unedited tree $\Phi_{\mathrm{rag}}(\mathcal{S})$, edited tree $\Phi_{\mathrm{rag}}'(\mathcal{S})$, selected ids $\mathcal{T}$, edit intent $I$
\ENSURE Merged tree $\Phi_{\mathrm{merged}}(\mathcal{S})$
\IF{$\mathcal{T}\neq\emptyset$ and $\Phi_{\mathrm{rag}}(\mathcal{S})$ contains no id in $\mathcal{T}$}
    \STATE \RETURN $\Phi(\mathcal{S})$
\ENDIF
\STATE Initialize $\Phi_{\mathrm{merged}}$ with edited root attributes if $\mathcal{T}=\emptyset$ or the root id is in $\mathcal{T}$; otherwise use original root attributes.
\FOR{each child id $u$ in the union of Original, Retrieved, and Edited children}
    \IF{$u$ exists in all three versions}
        \STATE Add $\mathrm{Merge}(\Phi_u, \Phi_{\mathrm{rag},u}, \Phi'_{\mathrm{rag},u}, \mathcal{T}, I)$.
    \ELSIF{$u$ exists in Edited and Original but not Retrieved}
        \STATE Add edited child after resolving identifier conflicts.
    \ELSIF{$u$ exists only in Edited}
        \IF{$\mathcal{T}=\emptyset$ or $I \in \{\mathrm{add}, \mathrm{replace}\}$}
            \STATE Add the edited child.
        \ENDIF
    \ELSIF{$u$ exists in Retrieved but not Edited}
        \IF{$I \in \{\mathrm{remove}, \mathrm{replace}\}$ and the retrieved subtree of $u$ contains an id in $\mathcal{T}$}
            \STATE Skip $u$ as an intended deletion.
        \ELSE
            \STATE Preserve the Original child if available; otherwise preserve the Retrieved child.
        \ENDIF
    \ELSIF{$u$ exists only in Original}
        \STATE Preserve the Original child.
    \ENDIF
\ENDFOR
\STATE \RETURN $\Phi_{\mathrm{merged}}(\mathcal{S})$
\end{algorithmic}
\end{algorithm}
\FloatBarrier

\noindent\textbf{Example.}
When editing a target object (e.g., a bed), a nearby contextual object (e.g., a desk) can appear in the retrieved partial DSL as an empty leaf. If we directly replace the subtree, this pruned contextual node would overwrite the fully-expanded version in the original scene, erasing its children (e.g., a computer). The three-way merge keeps the original desk subtree because the desk is context rather than a selected target, while still applying the bed's edited position and preserving its asset metadata across merge steps.

\subsection{Representation Scope}
Pure spatial hierarchies are less concise than procedural programs for repetitive patterns. We intentionally adopt a static declarative tree because it is retrieval-friendly: nodes can be indexed and edited directly without executing loops or resolving program state, which is important for localized editing. Control-flow abstractions can be layered on top by external template compilers that emit HDSL, preserving compatibility while keeping the core representation simple and reusable.

\section{Additional Qualitative Comparisons}

For large-room prompts, we raised Holodeck's maximum room-size limit to match our experimental setting. In our I-Design runs, complex layouts sometimes enter repeated layout-modification loops or recursive deadlocks, so we added forced termination and an upper bound on recursion depth. For consistent visual comparison, we also integrated a rendering module into I-Design and selected representative camera viewpoints with matched lighting. Additional HDSL generation examples are shown in~\cref{fig:comparison}.

\begin{figure}[!htbp]
    \centering
    \includegraphics[width=\columnwidth]{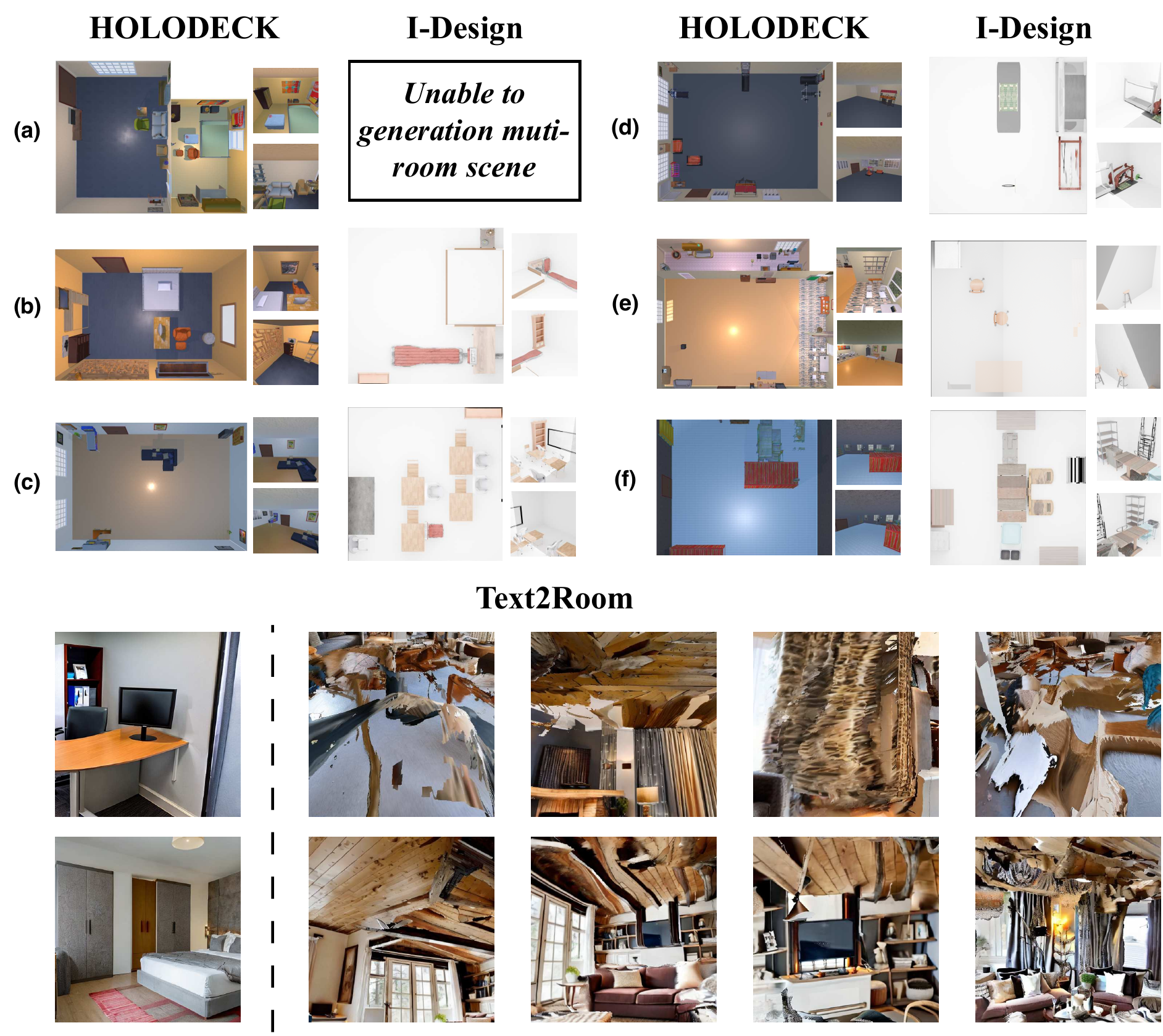}
    \caption{Additional baseline examples in sparse or open layouts. Holodeck tends to place objects near walls, I-Design relies on predefined room geometry and struggles with open-plan or multi-room settings, and Text2Room~\cite{hoellein2023text2room} degrades outside familiar image-generation distributions.}
    \label{fig:comparison_baseline}
\end{figure}

\noindent\textbf{Finer-grained instruction following.}
We provide an additional case showing that the LLM agent can handle finer-grained constraints in the prompt, including directions, relative ordering, and explicit object-count constraints inside each target scene configuration for testing.

\begin{figure}[!htbp]
    \centering
    \includegraphics[width=\columnwidth]{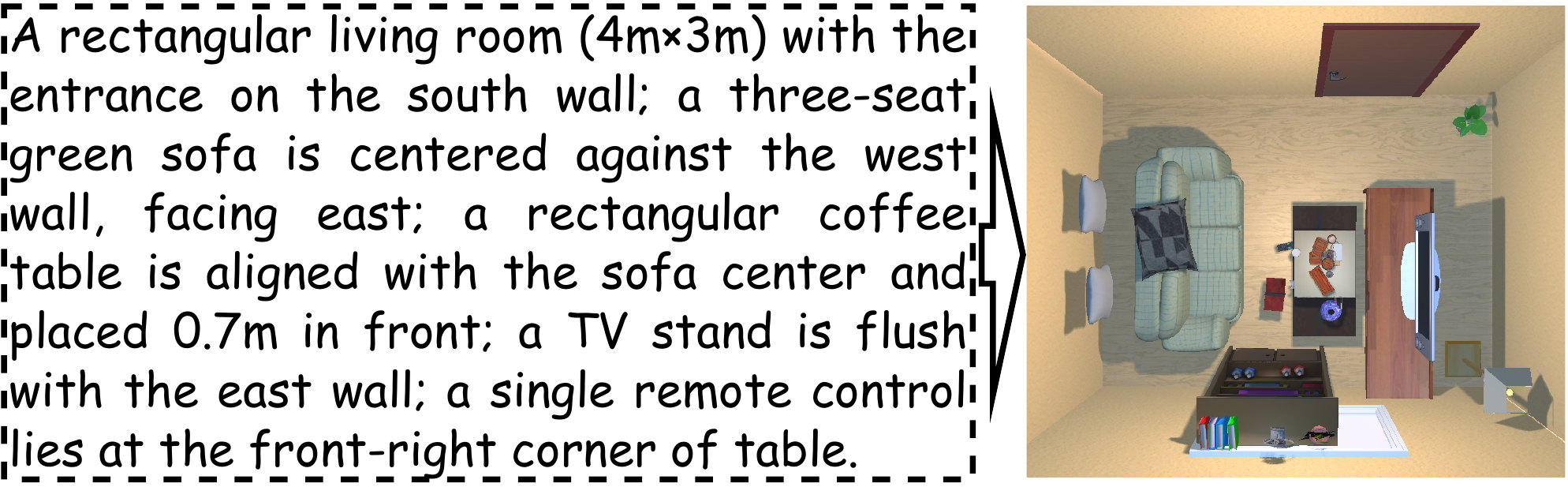}
    \caption{An example of finer-grained instruction following. The generation prompt is shown on the left and the resulting layout is shown on the right.}
    \label{fig:finer_grained_instruction_suppl}
\end{figure}

\FloatBarrier
\section{User Study Questionnaire}

The user study has two parts. For generation realism, 20 participants view three anonymized generated scenes, each shown from four camera views, and choose the most realistic scene. For editing, the same participants compare an original scene with the result of an instruction-based edit and rate compliance on a three-point scale: ``exactly matches," ``mostly matches," or ``does not match." Participants were general users, not members of the method team, and the study did not require professional 3D modeling experience from participants.

\begin{figure}[!htbp]
    \centering
    \includegraphics[width=0.8\linewidth]{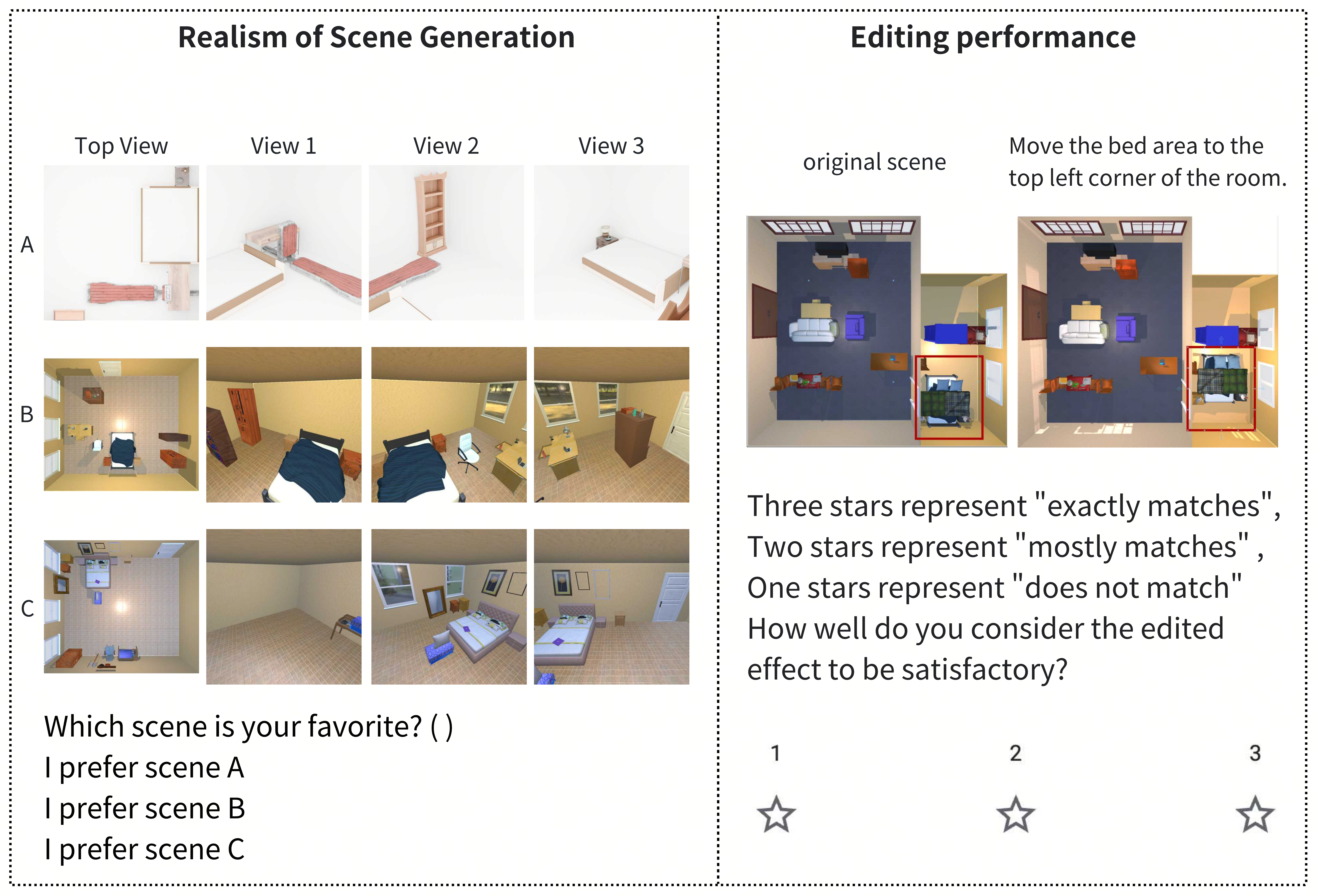}
    \caption{Sample questionnaire, consisting of two parts: generation realism (6 questions) and editing performance ratings (9 questions).}
    \label{fig:user_study_questionnaire}
\end{figure}

For the generation study, we use 5 representative scenes per category to capture structural diversity across varying room shapes and sizes while keeping human evaluation costs manageable. The display order of methods is randomized to mitigate bias, and participants do not see method names. We collected 120 votes for the generation evaluation and converted votes into Bradley--Terry preference scores for quantitative comparison. For editing, we collected 180 votes on the three-point compliance scale. The questionnaire does not ask participants to label error type, but the automatic edit logs show that non-exact HRAG cases are mostly small target-position misses after refinement rather than parser failures, object deletions, or metadata loss.

\FloatBarrier
\section{Prompt Templates}
The released implementation uses the templates in \texttt{hierarchical\_prompts.py}. For readability, we report the executable constraints and template variables here while omitting long few-shot examples.

\subsection{Generation Prompt}

\begin{tcolorbox}[mysinglecolumnbox]
\textbf{Input.} \texttt{INPUT\_CHAIN}, \texttt{INPUT\_LEAF}, and the user requirement. The model designs only the direct children of \texttt{INPUT\_LEAF}; valid descendants may be nested when they express physical support.

\textbf{Output.} Return exactly one XML-parsable HTML-like DSL fragment. The output must start with \texttt{<div} and end with \texttt{</div>}; no strategy text, markdown fence, heading, or explanation is allowed.

\textbf{Required attributes.} Every node includes \texttt{id}, \texttt{description}, \texttt{class}, \texttt{size}, \texttt{variance\_type}, \texttt{virtual}, \texttt{is\_leaf}, and inline \texttt{style}. The top-level node preserves the original \texttt{INPUT\_LEAF} attributes.

\textbf{Core rules.} Use \texttt{virtual=true} for zones and support containers, and \texttt{virtual=false} for objects grounded by retrieval. Parent-child nesting denotes support, e.g., desk-to-laptop and counter-to-sink are valid, while table-to-chair and room-to-sink are invalid. Coordinates use the parent frame with 1px = 1cm, and \texttt{transform: rotate(...deg)} controls top-view orientation.

\textbf{Spatial constraints.} Door spans are hard keep-clear regions and window spans are soft keep-clear regions. Do not emit clearance zones, aisles, corridors, walkable paths, empty areas, or keep-clear regions as DSL nodes. Avoid sibling overlap, keep wall-adjacent furniture facing the open room, and reduce optional large objects in compact bedrooms or kitchens before producing crowded layouts.

\textbf{Class constraints.} Do not create aggregate concrete classes such as \texttt{small\_items}, \texttt{small\_cooking\_items}, \texttt{accessories}, \texttt{decor}, or \texttt{miscellaneous\_items}; use concrete classes such as \texttt{pot}, \texttt{book}, \texttt{lamp}, \texttt{toaster}, or \texttt{coffee\_maker}.
\end{tcolorbox}

\subsection{Reflection Prompt}

\begin{tcolorbox}[mysinglecolumnbox]
Given \texttt{REJECT\_REASON} and the former DSL fragment, return a corrected XML-parsable fragment only. The prompt repeats the no-markdown and no-clearance-node rules. If the rejection is caused by overlap, crowding, or too many physical sibling pairs, the model is instructed to remove optional large objects rather than shifting the same crowded set. Compact-bedroom and compact-kitchen cases receive explicit pruning guidance.
\end{tcolorbox}

\subsection{Editing Prompt}

\begin{tcolorbox}[mysinglecolumnbox]
\textbf{Input.} The edit instruction, the retrieved HDSL context, and the root identifier of the provided scene.

\textbf{Output.} Return the full edited DSL fragment with the same root id as the provided scene; do not return only the edited child node. The output must be XML-parsable and contain no strategy text, markdown fence, heading, note, or explanation.

\textbf{Preservation rules.} Preserve existing node ids and unrelated sibling subtrees. For existing nodes that are not being replaced, keep \texttt{class}, \texttt{retrieved\_asset\_id}, \texttt{retrieved\_size}, \texttt{retrieved\_description}, and other asset metadata unchanged when present; update only the style, coordinates, rotation, or child set needed by the instruction.

\textbf{Layout rules.} Use only \texttt{<div>} tags with inline CSS. Each element includes \texttt{id}, \texttt{description}, \texttt{class}, \texttt{size}, \texttt{variance\_type}, \texttt{virtual}, and \texttt{is\_leaf}. The same support, generic-class, door/window, overlap, and clearance-node constraints used for generation also apply during editing.
\end{tcolorbox}

At runtime, the editor appends three short guards to this template: context siblings are preserved unless explicitly targeted, the returned fragment must keep the same root id, and an intent-specific suffix constrains add, remove, replace, rearrange, or modify instructions.

\end{document}